\documentclass[letterpaper, 10 pt, journal, twoside]{ieeetran}

\IEEEoverridecommandlockouts                              

\usepackage{times}
\usepackage{graphicx}
\usepackage{graphics} 
\usepackage[tight]{subfigure}
\usepackage{epsfig} 
\usepackage{mathptmx} 
\usepackage{times} 
\usepackage{amsmath} 
\usepackage{amssymb,amsthm,bm}  
\usepackage[ruled, linesnumbered, vlined]{algorithm2e}
\usepackage{floatrow} 
\usepackage{flushend}
\usepackage{cite}
\usepackage{mathtools}
\usepackage{soul}
\usepackage{xcolor,colortbl}
\usepackage{textcase}
\usepackage[tablename=TABLE]{caption}
\usepackage{url}

\definecolor{Gray}{gray}{0.85}
\newcolumntype{a}{>{\columncolor{Gray}}c}

\DeclareMathAlphabet{\mathcal}{OMS}{cmsy}{m}{n}

\begin{document}
\title{Polyline Generative Navigable Space Segmentation \\ for Autonomous Visual Navigation}

\author{Zheng Chen$^{1}$, Zhengming Ding$^{2}$, David J. Crandall$^{1}$, Lantao Liu$^{1}$
\thanks{Manuscript received: October, 17, 2022; Revised January, 06, 2023; Accepted January, 30, 2023.}
\thanks{This paper was recommended for publication by Editor Eric Marchand. Name upon evaluation of the Associate Editor and Reviewers' comments.
} 
\thanks{ 
$^{1}$Z. Chen, D. Crandall, L. Liu are with the Luddy School of Informatics, Computing, and Engineering  at Indiana University, Bloomington, IN 47408, USA. E-mail:
        {\tt\small \{zc11, djcran, lantao\}@iu.edu} 
        
$^{2}$Z. Ding is with the Department of Computer Science at Tulane University. Email: 
{\tt\small zding1@tulane.edu}
}%
\thanks{A preliminary version of this work appeared
as a poster in the 2021 NeurIPS Workshop on Machine Learning for Autonomous Driving.}
\thanks{This research is supported by NSF \#2006886 and ARL W911NF-
20-2-0099.}
\thanks{Digital Object Identifier (DOI): see top of this page.}
}

\markboth{IEEE Robotics and Automation Letters. Preprint Version. Accepted February, 2023}
{Chen \MakeLowercase{\textit{et al.}}: Polyline Generative Navigable Space Segmentation for Autonomous Visual Navigation} 

\maketitle

\begin{abstract}
Detecting navigable space is a fundamental capability for mobile robots navigating in unknown or unmapped environments. In this work, we treat  visual navigable space segmentation as a scene decomposition problem and propose  \textbf{P}olyline \textbf{S}egmentation \textbf{V}ariational autoencoder \textbf{Net}work (PSV-Net), a representation learning-based framework for learning the navigable space segmentation in a self-supervised manner. Current segmentation techniques heavily rely on fully-supervised learning strategies which demand a large amount of pixel-level annotated images. In this work, we propose a framework leveraging a  Variational AutoEncoder~(VAE) and an AutoEncoder~(AE) to learn a polyline representation that compactly outlines the desired navigable space boundary. Through extensive experiments, we  validate that the proposed PSV-Net can learn the visual navigable space with no or few labels, producing an  accuracy comparable to  fully-supervised state-of-the-art methods that use all available labels. In addition, we show that integrating the proposed navigable space segmentation model with a visual planner  can achieve efficient mapless navigation in real environments. 
\end{abstract}


\section{Introduction}
\IEEEPARstart{F}{or} mobile robots to be able to navigate unknown spaces, it is crucial to 
understand the traversability of complex environments that consist of cluttered objects. The goal is to construct collision-free traversable space, which we term as {\em navigable space}. 
If cameras are used to perceive the environment, a typical way to identify navigable space is through image segmentation by leveraging deep neural networks  to perform multi-class~\cite{yang2018denseaspp} or binary-class~\cite{fan2020sne} segmentation of images. 
The present work takes a binary-class segmentation approach in which the robot needs to distinguish navigable space from non-navigable space in the robot's perceived first-person-view (FPV) images.

However, most existing deep neural network-based methods are developed on top of a fully-supervised learning paradigm and rely on annotated datasets such as Cityscapes~\cite{cordts2015cityscapes} or KITTI~\cite{geiger2013vision}. These datasets usually contain an immense number of pixel-level annotated segmented images. Collecting and annotating such data for robotic applications in novel environments is prohibitively costly and time-consuming. 

\begin{figure} 
{
\centering
  {\includegraphics[width=0.98\linewidth]{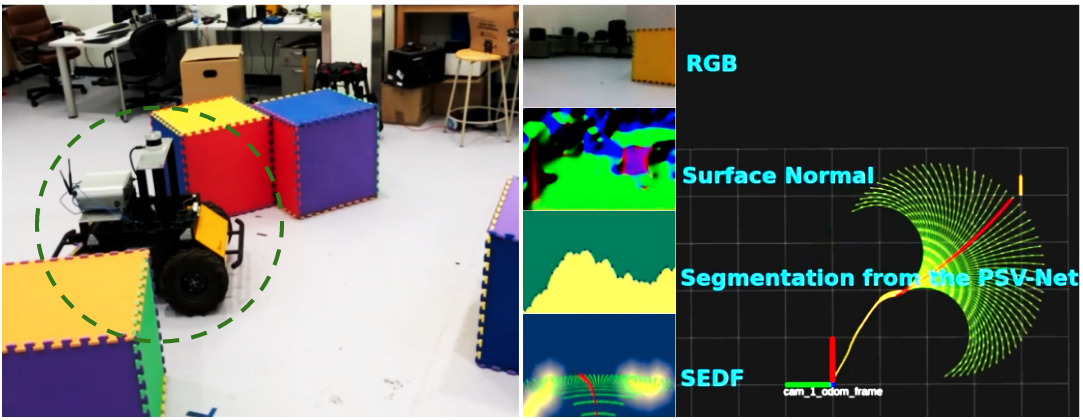}}
\caption{\small Navigation in a real cluttered environment with the proposed PSV-Net. The ground robot is circled in the left image. The right image shows  a motion primitive-based visual planner \cite{chen2022cali} that is used for navigation. The yellow arrow indicates the goal, the green paths are motion primitives, the red path is the optimal one, and the yellow path is the historically traversed path. SEDF represents the Scaled Euclidean Distance Field; see \cite{chen2022cali}. \vspace{-15pt}
} 
\label{fig:real_exp1}  
} 
\end{figure}

To overcome the limitation of fully-supervised learning and pave a path for mobile robot navigation, we propose   a self-supervised learning method by treating  binary navigable space segmentation as a {\em scene decomposition} problem, and the training signals come from certain visual inputs such as surface normal images. It has been demonstrated in \cite{burgess2019monet, greff2019multi} that  scene decomposition can be solved in an unsupervised fashion with Variational AutoEncoders~(VAEs). In contrast to semantic segmentation by supervised learning where the model is trained by human-annotated pixel-wise labels, scene decomposition attempts to learn the compositional nature from  visual information alone without explicit annotations.

\begin{figure*} [t] \vspace{8pt}
\floatbox[{\capbeside\thisfloatsetup{capbesideposition={right,top},capbesidewidth=6.0cm}}]{figure}[\FBwidth]
{\label{fig:turbine1}\includegraphics[height=2.2in]{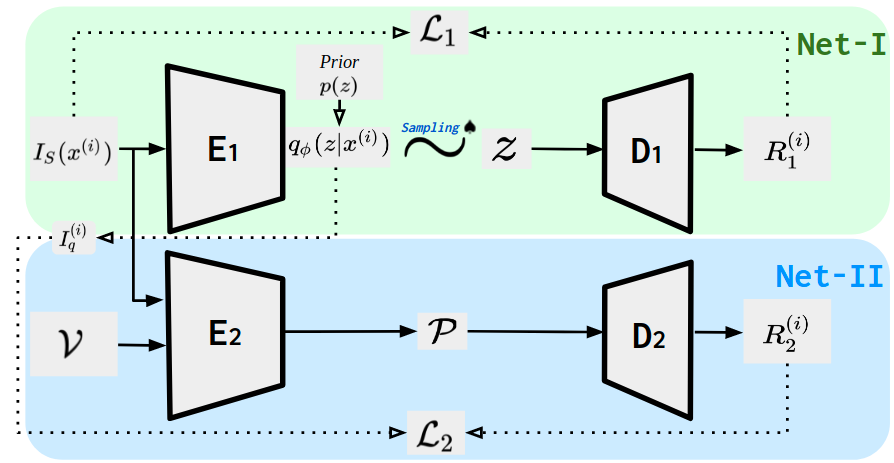}}
{\caption{\small  
Framework overview. $I_{S}$, $E_1/E_2$, and $D_1/D_2$ are the input surface normal image, encoders, and decoders for both nets. \textbf{Net-I}: $q_{\phi}(z|x^{(i)})$ represents the latent categorical distribution from which a segmentation sample $z$ is sampled using $\spadesuit$ -- the Gumbel-Softmax sampler for discrete distributions proposed in \cite{jang2016categorical}. Prior I is to regularize the ``shape'' of the predicted latent distribution. $R_1^{(i)}$ is the reconstructed input and $\mathcal{L}_{1}$ is the reconstruction loss between $I_{S}$ and $R_1^{(i)}$. \textbf{Net-II}: $\mathcal{V}$ are the initial vertices for polyline representation. $\mathcal{P}$ is a set of predicted vertices from $E_2$. $R_2^{(i)}$ is the reconstructed image and $\mathcal{L}_{2}$ is the reconstruction between $R_2^{(i)}$ and the latent image $I_q$, converted from the latent distribution $q_{\phi}$.
} 
\label{fig:nets}
} \vspace{-15pt}
\end{figure*} 

Most (if not all) existing VAE-based scene decomposition and representation learning methods use pixel-wise learning in which the value of every pixel is predicted. 
However, pixel-wise methods prone to generate segmentation with noises, e.g., pixel islands or scatters, which can inevitably affect downstream decision-making in the planning module, leading to the whole navigation system being vulnerable to unsafe/inefficient decisions. In addition, it is also expensive to adopt a post-processing module to smooth out possible pixel noises. An intuitive way to solve issues of pixel-wise methods is to directly learn the desired boundary for downstream planning tasks. The boundary could be represented by polylines/splines (represented by vertices or control points)~\cite{wang2019object,peng2020deep}, which are naturally robust to  noise in images. 

The proposed PSV-Net consists of two networks, Net-I (a VAE) and Net-II (an AE). During training, the goal of Net-I is to learn pseudo labels from surface normals by using categorical distributions as the latent representation. Using the supervision signals from  Net-I,    Net-II learns a set of vertices which describe the location of the navigable space boundary. We  show that our boundary-based PSV-Net  achieves comparable segmentation results to  fully-supervised learning-based state-of-the-art methods. To demonstrate the  efficacy of our PSV-Net on real navigation tasks, we also integrate the proposed segmentation model with a visual planner to guide the robot to move without collision (see Fig.~\ref{fig:real_exp1}).

\section{Related Work}
Traditional image segmentation methods with no supervision focus on crafting features and energy functions to define desired objectives. For example, active contour-based models \cite{kass1988snakes} optimize over a polygon (represented by vertices) by means of energy minimization based on both the image features and  shape priors such as boundary continuity and smoothness. However, active contours lack flexibility and heavily rely on low-level image features and global parameterization of priors~\cite{marcos2018learning}. Recently, deep active contour-based models have been proposed~\cite{marcos2018learning}, but these  methods require ground-truth contour vertices and thus belong to the supervised learning paradigm. Another line of research uses adversarial approaches for unsupervised segmentation, e.g., work in \cite{chen2019unsupervised} explores the idea of changing the textures or colors of objects without changing the overall distribution of the dataset, and proposes an adversarial architecture to segment a foreground object in each image. Although the adversarial methods show impressive results, they suffer from instabilities in training. The proposed technique in this paper is close in spirit to some work in scene decomposition~\cite{burgess2019monet} where the goal is to decompose scenes into objects in an unsupervised fashion with a generative model. 

Our proposed framework is built upon VAEs,
which are often used for 
representation learning
\cite{tschannen2018recent,bengio2013representation}   to learn useful representations of data with little or no supervision. The trained generative model (decoder) can generate new images from any random sample in the learned latent distribution. Many variants of VAEs have been proposed following \cite{kingma2013auto}, where the latent representation is described as a continuous distribution. In this paper, to learn the navigable space, instead, we use a discrete distribution as the latent representation.

In addition, the value of boundaries in image segmentation has been shown in a large amount of previous literature. The  polygon/spline representation is used in \cite{castrejon2017annotating, acuna2018efficient, ling2019fast, peng2020deep, Liang_2020_CVPR} to achieve fast and potentially interactive instance segmentation. Acuna {\em et al} propose a new approach to learn sharper and more accurate semantic boundaries \cite{acuna2019Steal}. By treating  boundary detection as the dual task of semantic segmentation, a new loss function with a boundary consistency constraint to improve boundary pixel accuracy for semantic segmentation is designed \cite{Zhen_2020_CVPR}. The work in \cite{Marin_2019_ICCV} proposes a content-adaptive downsampling technique that learns to favor sampling locations near semantic boundaries of target classes. By doing so, segmentation accuracy and computational efficiency can be balanced. Although the above methods make use of boundaries to improve  performance, they all require ground truth labels for training --- an important limitation that we aim to overcome in this work. 

\section{Methodology}

The proposed PSV-Net consists of two sub-nets, as shown in Fig.~\ref{fig:nets}. Our intuition for this two-net architecture is based on a coarse-to-fine process, where Net-I learns a rough pixel-wise pseudo label of navigable space by taking advantage of generative modeling, while Net-II generates a refined version of the prediction from Net-I using a set of vertices.

\subsection{Net-I}
The goal of Net-I is to learn a pixel-wise pseudo label from the input surface normal image.
A standard VAE aims to learn a generative model which can synthesize new data from a random sample in a specified latent space. We want to obtain the parameters of the generative model by maximizing the data marginal likelihood $\log p_{\theta}(\mathbf{x}^{(i)})$, where $\mathbf{x}^{(i)}$ is a data point in our training dataset $\left \{ \mathbf{x}^{(i)} \right \}_{i=1}^N$. Using Bayes' rule, the likelihood can be written,
\begin{equation}
\label{eq:marginal_likelihood}
    \log p_{\theta}(\mathbf{x}^{(i)}) = \log \sum\nolimits_{\mathbf{z}} \Bigg[ \Bigg. p_{\theta}(\mathbf{x}^{(i)}|\mathbf{z})p_{\theta}(\mathbf{z})\Bigg. \Bigg],
\end{equation}
where $p_{\theta}(\mathbf{z})$ is the prior distribution of the latent representation and $p_{\theta}(\mathbf{x}^{(i)}|\mathbf{z})$ is the generative probability distribution of the
reconstructed input given the latent representation. We can use a neural network to approximate $p_{\theta}(\mathbf{x}^{(i)}|\mathbf{z})$ where $\theta$ can be thought as parameters of the network, but we cannot perform the sum operation over $\mathbf{z}$, hence Eq.~(\ref{eq:marginal_likelihood}) is computationally intractable. 

An alternative  is to find the lower bound of Eq.~(\ref{eq:marginal_likelihood}); to do this,  
we can use an encoder to approximate the true posterior of the latent variable $p_{\theta}(\mathbf{z}|\mathbf{x}^{(i)})$. We denote the encoder (another neural network) as $q_{\phi}(\mathbf{z}|\mathbf{x}^{(i)})$, where $\phi$ are parameters of the encoder network. Then we can derive the lower bound of Eq.~(\ref{eq:marginal_likelihood}) as \cite{kingma2013auto},
\begin{equation}
    \label{eq:elbo}
    \begin{aligned}
        \log p_{\theta}(\mathbf{x}^{(i)}) = &\mathbb{E}_{\mathbf{z}} \Bigg[ \Bigg.
        \log p_{\theta} (\mathbf{x}^{(i)}|\mathbf{z}) \Bigg. \Bigg] - \mathbb{KL}(q_{\phi}(\mathbf{z}|\mathbf{x}^{(i)}) || p_{\theta}(\mathbf{z}))+ \\
        &\mathbb{KL}(q_{\phi}(\mathbf{z}|\mathbf{x}^{(i)}) || p_{\theta}(\mathbf{z}|\mathbf{x}^{(i)})).
    \end{aligned}
\end{equation}

\begin{figure} \vspace{7pt}
\centering
  {\includegraphics[width=\linewidth]{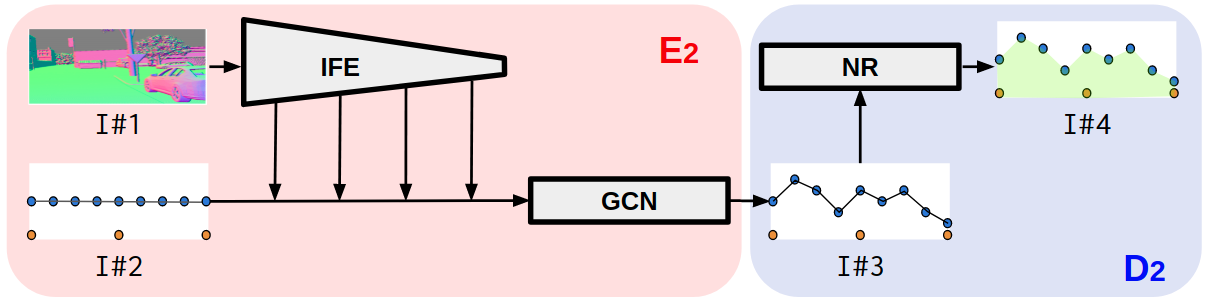}}
\caption{\small Structure of Net-II. $I\#1$: Surface normal image, $I\#2$: Initial vertices, $I\#3$: Newly predicted vertices, and $I\#4$: Rendered output. $E_2$: encoder of Net-II. IFE: Image Feature Extraction net. GCN: Graph Convolutional Network. $D_2$: decoder of Net-II. NR: Neural Rendering module. In both blocks, blue points are vertices of polylines while  orange points are auxiliary points on image bottom boundary for the convenience of neural rendering. \vspace{-15pt} 
} 
\label{fig:net2}  
\end{figure}

To perform  segmentation, we assume $p_{\theta}(\mathbf{x}^{(i)} | \mathbf{z}) = \prod_{j=1}^J \mathcal{N} (\mathbf{x}^{(i)}; R_1^{(i)}, \sigma^2)$, $q_{\phi}(\mathbf{z}|\mathbf{x}^{(i)}) = \prod_{j=1}^J Cat(z_j|\mathbf{x}^{(i)}, \phi)$, and the prior $p_{\theta}(\mathbf{z}) = \prod_{j=1}^J Cat(z_j)$, where $J$ is the number of pixels in the input image (same meaning in the following equations). Then only the last term $\mathbb{KL}(q_{\phi}(\mathbf{z}|\mathbf{x}^{(i)}) || p_{\theta}(\mathbf{z}|\mathbf{x}^{(i)})) $ of Eq.~(\ref{eq:elbo}) is unknown. Fortunately, we know that $\mathbb{KL}(p||q)\geq 0$ is always true for any two distributions $p$ and $q$. Therefore, 
\begin{equation}
    \label{eq:lower_bound}
    \log p_{\theta}(\mathbf{x}^{(i)})\geq \mathbb{E}_{\mathbf{z}} \Bigg[ \Bigg.
    \log p_{\theta} (\mathbf{x}^{(i)}|\mathbf{z})\Bigg. \Bigg] - \mathbb{KL}(q_{\phi}(\mathbf{z}|\mathbf{x}^{(i)}) || p_{\theta}(\mathbf{z})),
\end{equation}
where the term on the right side of the inequality is called the \textit{Evidence Lower BOund}~(ELBO) of the data marginal likelihood. Maximizing the ELBO is equivalent to maximizing the data marginal likelihood. By using Monte Carlo sampling, the total loss to be \textit{minimized} over the whole training dataset is:
\begin{equation}
    \label{eq:loss_1}
    \begin{aligned}
        \mathcal{L}_{I} = &\sum\nolimits_{i=1}^N  \Bigg[ \Bigg. \mathbb{KL}(q_{\phi}(\mathbf{z}|\mathbf{x}^{(i)}) || p_{\theta}(\mathbf{z})) +\\ &\frac{1}{2KJ\sigma^2}\sum\nolimits_{k=1}^K \left ( \left \| \mathbf{x}^{(i)} - R_{1k}^{(i)} \right \|_2^2 + \frac{J}{2} \log \sigma^2 \right ) \Bigg. \Bigg],
    \end{aligned}
\end{equation}
where $J$ is the total number of pixels, $N$ is the number of images in our training dataset, and $K$ is the number of samples in Monte Carlo sampling. $R_{1k}^{i}$ is the reconstructed input from the $k^{th}$ sampled $z$ in Monte Carlo sampling. The reconstruction loss $\mathcal{L}_1$ in Fig.~\ref{fig:nets} is equivalent to the term  $\frac{1}{2K\sigma^2}\sum\nolimits_{k=1}^K \left \| \mathbf{x}^{(i)} - R_1^{(i)} \right \|_2^2$.

For Net-I, we use a U-Net \cite{ronneberger2015u} as the $E_1$ to approximate the latent true posterior, and one convolutional layer as the $D_1$ to reconstruct an image from the latent code. The output of  $E_1$, $q_{\phi}(\mathbf{z}|\mathbf{x}^{(i)})$, represents the categorical distribution, from which we can obtain a pesudo label by computing $I_q^{(i)} = \arg \max_{c} q_{\phi}(\mathbf{z}|\mathbf{x}^{(i)})$, where $c$ is the channel number of the output in $E_1$. We use $I_q^{(i)}$ as the pseudo label to supervise the training of  Net-II, which we now describe.

\subsection{Net-II}
With the supervision signals from  Net-I,  Net-II aims to learn a set of vertices which describe the location of the navigable space boundary. Net-II is an autoencoder, where the encoder manipulates the coordinates of a set of vertices (from $\mathcal{V}$ to $\mathcal{P}$ in Fig. \ref{fig:nets}). Then the predicted vertices $\mathcal{P}$ are used to render a reconstructed image $R_2^{(i)}$ by the decoder $D_2$. The training of the Net-II is supervised by the pseudo label $I_q^{(i)}$ from  Net-I. The overall objective of Net-II, $\mathcal{L}_{II}$, is equal to the reconstruction loss $\mathcal{L}_2$ (see Fig. \ref{fig:nets}), which consists of a mean squared error and an appearance matching error~\cite{guizilini20203d} estimated using the Structural Similarity Index Measure (SSIM),
\begin{equation}
    \label{eq:loss_2_new}
    \mathcal{L}_{II} = \mathcal{L}_2 = \lambda_1 \frac{1 - SSIM(I_q^{(i)}, R_2^{(i)})}{2} + \lambda_2 \frac{1}{J}\left \|  I_q^{(i)} - R_2^{(i)}\right \|^2_2,
\end{equation}
where $J$ is the same as in Eq.~(\ref{eq:loss_1}), $\lambda_1$ and $\lambda_2$ are weights, and $\lambda_1 + \lambda_2 = 1$.

We use a hybrid net consisting of a convolutional neural network and a graph convolutional network as the encoder (see the \textit{left block}, $E_2$ in Fig.~\ref{fig:net2}), and a neural rendering module \cite{kato2018neural} as the decoder (see the \textit{right block}, $D_2$ in Fig.~\ref{fig:net2}). Specific Net-II components (shown in Fig. \ref{fig:net2}) are described as follows. 

\begin{figure} \vspace{7pt}
{
\centering
  {\includegraphics[width=\linewidth]{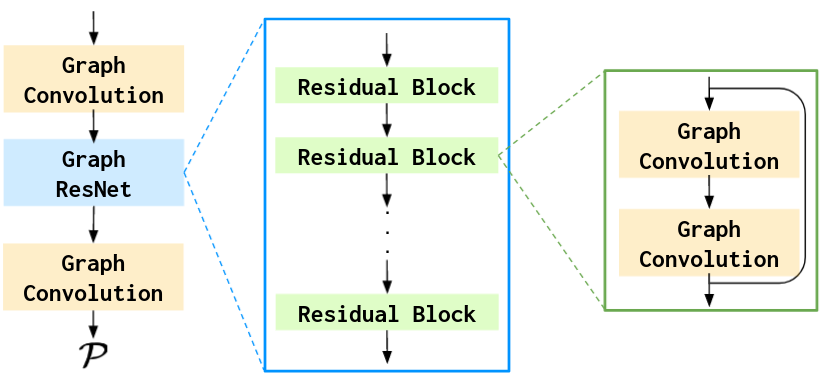} 
  } 
\caption{\small GCN module in the \textit{left block} of Fig.~\ref{fig:net2}. \vspace{-15pt}
}  
\label{fig:gcn} 
}
\end{figure}

\textbf{Image Feature Extraction (IFE):} 
The IFE module is adopted to generate feature maps at different resolutions, where different layers have different resolutions. The feature maps are then used as features for vertices in the GCN module (will be discussed next), where we use the coordinates of the vertices to index locations of the features.  Bilinear interpolation is applied to obtain the coordinates in different layers. The features from the IFE module give the downstream network more information regarding the vertices, such that the vertices' shift prediction can be well supported by visual clues. 

\begin{figure} \vspace{7pt}
{
  \centering
    \subfigure[]
  	{\label{fig:nr_1}\includegraphics[width=0.31\linewidth]{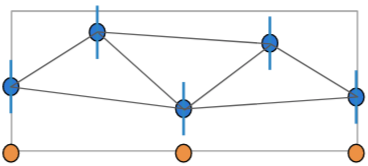}}
  	\subfigure[]
  	{\label{fig:nr_2}\includegraphics[width=0.31\linewidth]{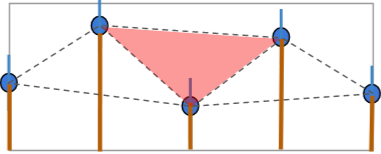}} 
  	\subfigure[]
  	{\label{fig:nr_3}\includegraphics[width=0.31\linewidth]{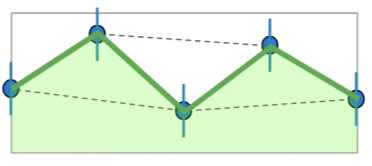}} 
  \caption{\small Triangle selection for neural rendering. (a) Auxiliary points (orange), example predicted vertices (blue), and constructed triangles (black lines, only those formed by boundary points are considered in the selection step). (b) Orange lines connect predicted vertices with corresponding projected vertices on the image bottom border. Only those triangles which do not intersect with any of orange lines are ignored (e.g., the red one) to render. (c) Desired navigable space from neural rendering is marked as green region. 
  }
\label{fig:nr}  
}
\end{figure}

\textbf{Graph Convolution Network (GCN):}
We construct a graph with each node as the concatenation of the extracted image features and their coordinates. Our GCN structure (see Fig.~\ref{fig:gcn}) is inspired by the network structure proposed in \cite{wang2018pixel2mesh, ling2019fast}.  
The fundamental layer of the GCN module is the graph convolutional layer. We denote a graph as $\mathcal{G}=\left \{ \mathcal{V}, \mathcal{E}, \mathcal{F} \right \}$, where $\mathcal{V} = \left \{ v_i \right \}_{i=1}^{T}$ denotes the node set, 
$\mathcal{E}=\left \{ e_j \right \}_{j=1}^{M}$ represents the edge set, and $\mathcal{F}=\left \{ f_i \right \}_{i=1}^{T}$ is the feature vector set for nodes in the graph. $T$ and $M$ are the number of nodes and edges, respectively. Then a graph convolution layer is defined as,
\begin{equation}
    \label{eq:gcn_layer}
    f_i^{l+1} = \Gamma(w_0^l f_i^l + \sum\nolimits_{j\in Nei(i)} w_1^l f_j^l),
\end{equation}
where $f_i^{l+1}$ and $f_i^l$ are the feature vectors on vertex $i$ before and after the convolution, $Nei(i)$ are the neighboring nodes of $i$, and $\Gamma(\cdot)$ is the activation function. 

Note that the vertices are iteratively manipulated twice by the GCN, but the IFE only performs a single forward pass. In each manipulation, we use the updated coordinates to extract new features from the same set of feature maps.

\textbf{Neural Rendering (NR): }
To convert the vertices predicted from the encoder $E_2$ to a \textit{segmentation image} such that  Net-II can be trained using the pseudo label from  Net-I, we triangularize those vertices and select proper triangles for neural rendering. We first select three auxiliary vertices $\mathcal{V}^{'}$ on the bottom boundary of image (see orange points of the \textit{left} block of Fig.~\ref{fig:net2}). The auxiliary vertices $\mathcal{V}^{'}$ are fixed through the GCN module such that the input to $D_2$ is a new set of vertices consisting of $\mathcal{V}^{'}$ and $\mathcal{P}$: $\mathcal{P}+ = \mathcal{P}\cup \mathcal{V}^{'}$.
The motivation for the auxiliary points is that we need to construct a \textit{closed} area for neural rendering processing.
We then use Delaunay triangulation to construct triangles on $\mathcal{P}+$. Next we use neural rendering~\cite{kato2018neural} to render the constructed triangles into a segmentation output while keeping the whole pipeline differentiable. However, a discrepancy exists: Delaunay triangulation always returns a series of triangles forming a \textit{convex} hull but the shape of real navigable space is not necessarily always convex, so the NR module tends to generate more triangles than needed. We propose to use a simple triangles selection method (see Fig.~\ref{fig:nr}) to filter out unnecessary triangles.

\begin{figure} \vspace{7pt}
{
\centering
  {\includegraphics[width=0.96\linewidth]{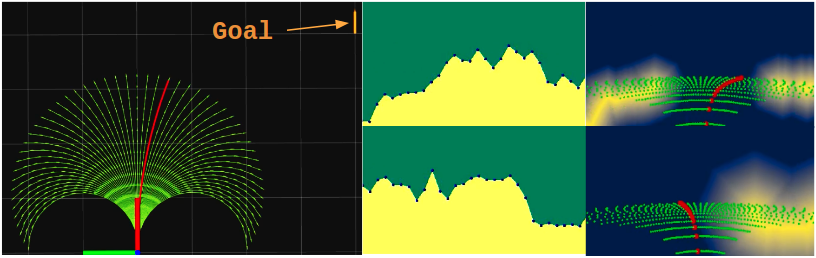}}
\caption{\small Integration of the visual planner \cite{chen2022cali} with segmentation model. \textit{Left:} Green curves: motion primitives in map space, Red curve: the optimal primitive, and Yellow arrow: Goal. \textit{Middle:} Segmentation images at two time steps during navigation. \textit{Right:} Scaled Euclidean Distance Field based on the segmentation images. The 2D projections of motion primitives are also visualized.
} 
\label{fig:primitive}  
}
\end{figure}

\subsection{Visual Planning}
To make the proposed model effective in real navigation, we integrate the PSV-Net with an image-based planner \cite{chen2022cali}, which projects a library of motion primitives from map space (see the left image of Fig. \ref{fig:primitive}) to image space. This allows us to only use the binary segmentation images to make the decision on which primitive is optimal to track, leading to a \textit{mapless} navigation system.

Specifically, first we convert the binary segmentation to a Scaled Euclidean Distance Field (SEDF); see two examples in right images of Fig. \ref{fig:primitive}. Values in the SEDF represent costs of colliding with obstacle boundaries. Next, we compute a library of motion primitives~\cite{howard2007optimal, howard2008state} $\mathcal{M} = \left \{ \mathbf{p}_1, \mathbf{p}_2, \cdots, \mathbf{p}_n \right \}$ projected from map space to SEDF.
Then we compute the navigation cost function for each primitive based on the evaluation on the SEDF and target progress (a distance measure from robot's current pose to the goal pose). Finally we select the primitive with the minimal cost to execute. The trajectory selection problem can be defined as,
\begin{equation}
    \label{eq:traj_selection}
    \mathbf{p}_{optimal} = \underset{\mathbf{p}}{\text{argmin}}~ w_1 \cdot C_{c}(\mathbf{p}) + w_2 \cdot C_{t}(\mathbf{p}),
\end{equation}
where $C_{c}(\mathbf{p}) = \sum_j^m c_c^j$ and $C_{t}(\mathbf{p}) = \sum_j^m c_t^j$ are the collision cost and target cost of one primitive $\mathbf{p}$, and $w_1$, $w_2$ are corresponding weights, respectively. Readers can find more details about the SEDF and how to compute target progress in \cite{chen2022cali}.

\section{Experiments}

\subsection{Baselines and Evaluation Metrics}
\label{sec:baselines_and_metrics}
To evaluate the proposed method, we compare against two baselines, SNE-RoadSeg \cite{fan2020sne} and OFF-Net \cite{min2022orfd}, both of which are pixel-wise, state-of-the-art methods for binary navigable space segmentation. We conduct extensive qualitative and quantitative experiments on multiple datasets. Note that our method takes depth/surface images as input. For fair comparison, all methods in our experiments require that at least depth images are provided. We use the mean of Intersection over Union (mIoU) as the metric for reporting quantitative results. The definition of IoU is: $\mathbf{IoU} = \frac{n_{tp}}{n_{tp} + n_{fp} + n_{fn}}$, where $n_{tp}, n_{tn}, n_{fp}$ and $n_{fn}$ are true positive, true negative, false positive, and false negative, respectively.

\begin{figure*} \vspace{7pt}
{
  \centering
    \subfigure[]
{\label{fig:kitti_qual}\includegraphics[width=0.487\linewidth]{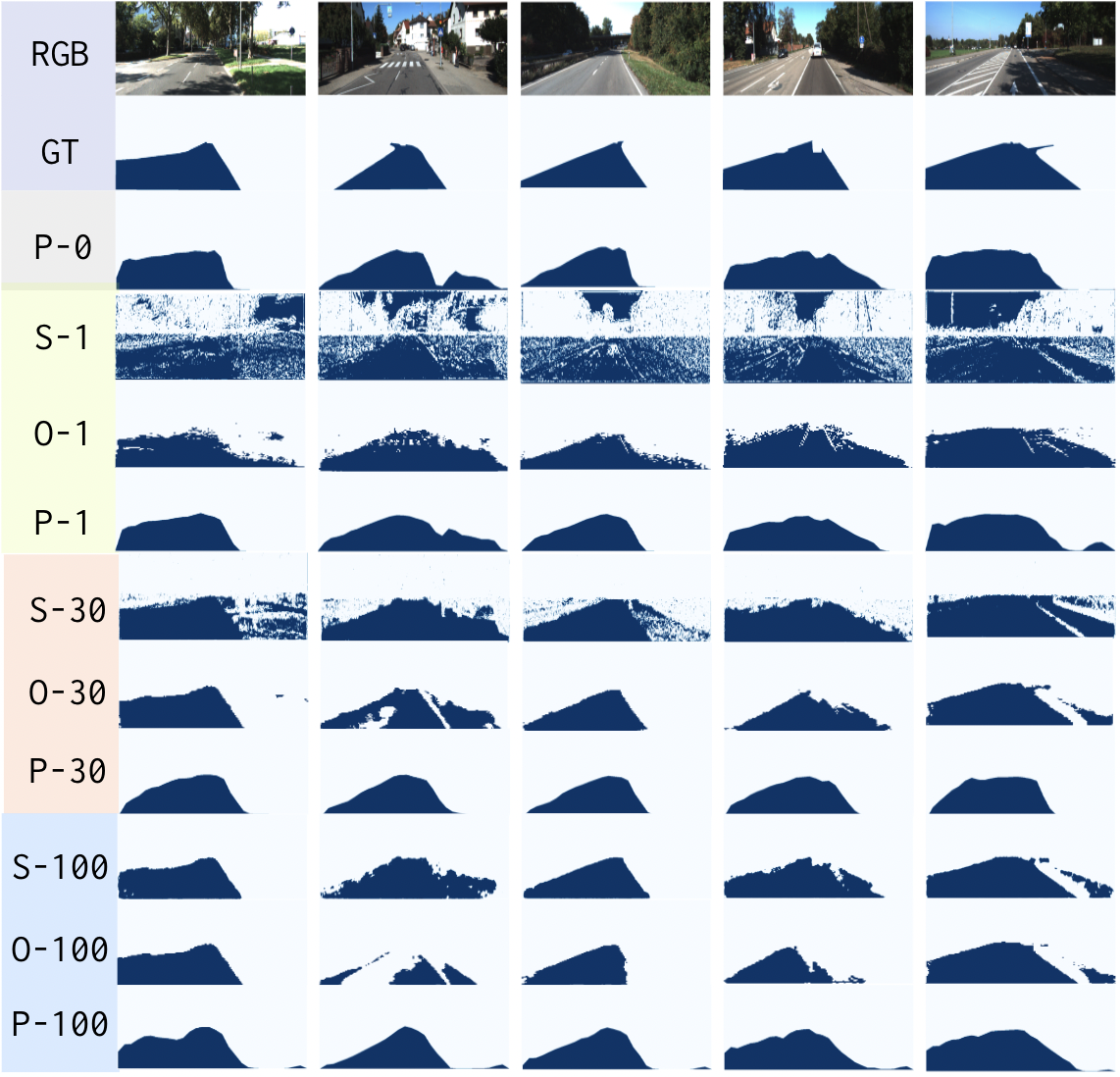}}
  	\subfigure[]
{\label{fig:city_qual}\includegraphics[width=0.49\linewidth]{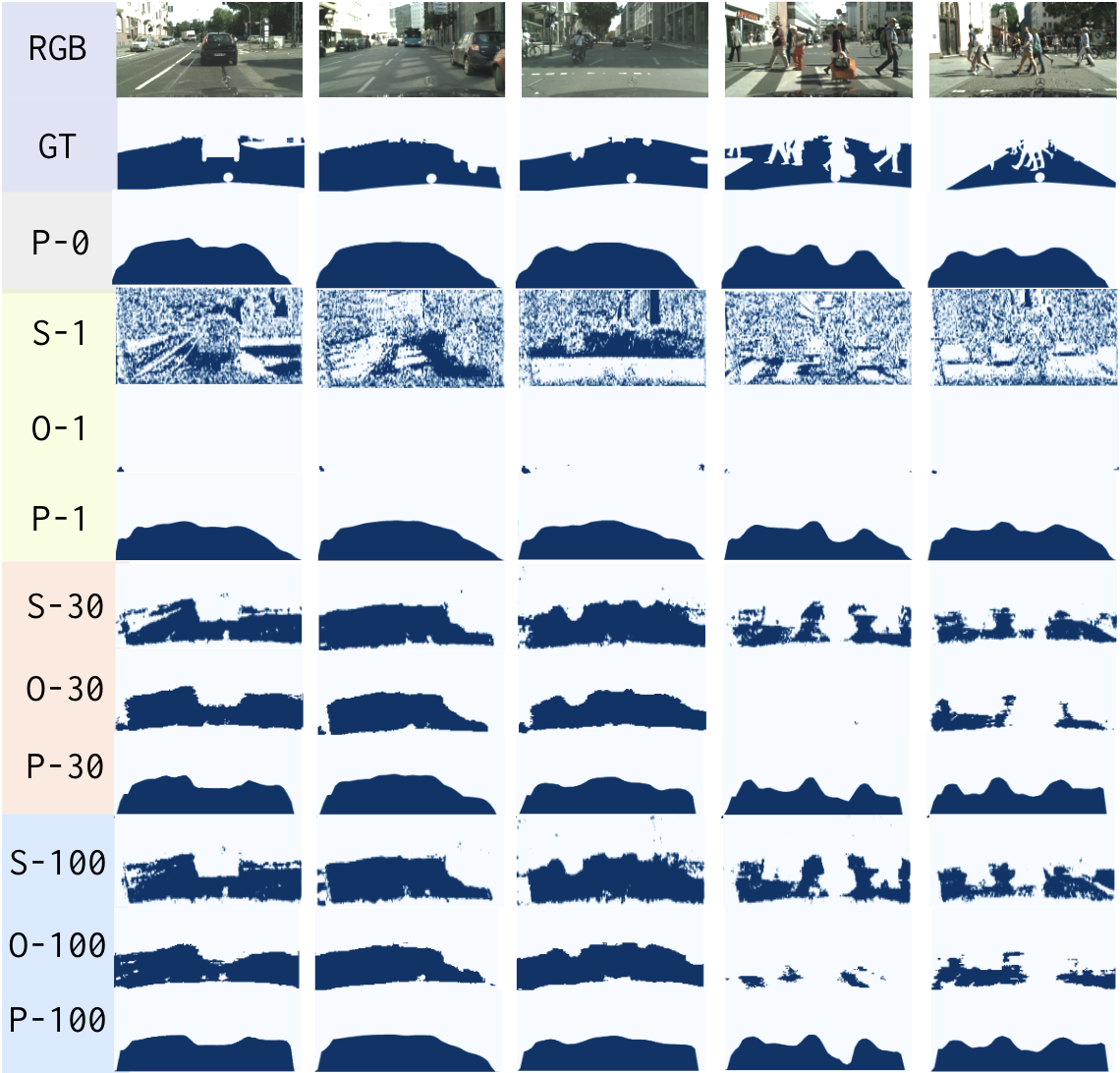}} 
  \caption{\small Qualitative results on (a) KITTI and (b) Cityscapes. First row: RGB images; Second row: ground-truth navigable space segmentation. Labels for each row of prediction results (from the third row to the last row) have the form S/O/P-$\#$, where S represents the baseline SNE-RoadSeg~\cite{fan2020sne}, O represents the baseline OFF-Net \cite{min2022orfd}, P represents the proposed method PSV-Net, and $\#$ represents the percentage of the  ground truth labels used. The baseline methods will not work if no ground truth labels are provided, so S-0 and O-0 are not available.
  } 
\label{fig:quals}  
}
\end{figure*}

\subsection{Datasets}
We evaluate the proposed method on several different datasets, including the standard KITTI road dataset~\cite{fritsch2013new}, the Cityscapes dataset \cite{cordts2015cityscapes}, and the ORFD dataset for off-road environments \cite{min2022orfd}.

The original KITTI road segmentation dataset \cite{cordts2015cityscapes} is specifically designed for binary navigable space segmentation in city-like environments. Since our proposed method learns  segmentation from images in a self-supervised manner, we can only learn free space from  images where geometric boundaries are consistent with human-defined labels. Otherwise, the self-learned segmentation will be different from the ground truth labels (but still reasonable), causing confusion for evaluation. For example, \textit{sidewalk} sometimes shares the same geometric plane with  \textit{road}, meaning both classes can be geometrically navigable, but in human-predefined labels only the class  \textit{road} is marked as navigable.  
Therefore, throughout the experiments (including the baseline experiments), we only use the images starting with \textit{um} and \textit{umm} with corresponding \textit{road} (instead of \textit{lane}) ground truth labels in training data. We convert the LiDAR point clouds to depth images and use the surface normal estimation (SNE) module proposed in \cite{fan2020sne} to compute surface normal images.

Cityscapes \cite{cordts2015cityscapes} is a standard semantic segmentation dataset in which pixel-wise semantic labels are provided for images across multiple cities. We treat the \textit{road} label as navigable while others as non-navigable, thus creating binary ground truth images.  Note that we apply no data selection to the Cityscapes data as there is no way to distinguish images which have inconsistency between the geometry and the semantic label (of the \textit{road}). This inevitably causes us to underestimate the performance of our proposed method. We use the provided disparity images to compute depth images and then obtain normal images using SNE in \cite{fan2020sne}.

ORFD \cite{min2022orfd} is a newly-released binary segmentation dataset for various off-road environments, e.g., farmland, grassland, woodland, countryside, and snowland. The data in ORFD have a notably different distribution than images in KITTI or Cityscapes, so we can test  generalization capability during adaptation from one dataset to another, e.g., KITTI $\rightarrow$ ORFD (meaning the model is trained using KITTI  but tested on ORFD).

To further validate  our proposed method on various data, we also build a small dataset which is collected in indoor environments. We name the data we collect as INDOOR dataset. Examples of the indoor data can be seen in Fig. \ref{fig:indoor_data}. Note that in INDOOR, only RGB and depth images are collected. The surface normal images are computed in the same way as \cite{fan2020sne}. The INDOOR dataset has no ground truth labels.

\subsection{Implementation Details}
To implement  PSV-Net as illustrated in Fig.~\ref{fig:nets}, 
$E_1$ is a U-Net~\cite{ronneberger2015u} with 2D convolution kernels and $D_1$ is a simple fully convolutional network. Similar to \cite{wang2018pixel2mesh}, IFE in $E_2$ is a VGG-16 \cite{simonyan2014very} network up to layer \textit{conv5\_4} and we use the concatenation of features from layer \textit{conv2\_3, conv3\_3, conv4\_3} and \textit{conv5\_4} as the visual features of the nodes in the graph. The GCN in $E_2$ is based on \cite{wang2018pixel2mesh, ling2019fast}. We use the pretrained mesh renderer proposed in \cite{kato2018neural} as  $D_2$~(NR). We fix the parameters of $D_2$, and only parameters of $E_1$, $D_1$, and $E_2$ are updated during training. We use the {\em Adam optimizer} for both networks and set the learning rate $lr_1 = 1e-3$ for  Net-I and $lr_2 = 1e-4$ for  Net-II. The total number of training epochs is 15 and we fix the weights of Net-I after epoch 3. We set the weights in $\mathcal{L}_2$~(Eq.~(\ref{eq:loss_2_new})) as $\lambda_1 = 0.8$ and $\lambda_2 = 0.2$. We compare to the most recent baseline \cite{fan2020sne}, a fully-supervised state-of-the-art method for predicting navigable space. To permit a fair comparison, we use the same batch size, 1, for both methods. We implement the whole framework using PyTorch~\cite{paszke2019pytorch} and conduct all experiments on a single Nvidia Geforce RTX 2080 Super GPU.

\subsection{Comparison for KITTI} 
A qualitative comparison of different methods on KITTI  is shown in Fig.~\ref{fig:kitti_qual}. Predictions are reported with 4 different percentages of ground truth labels ($0\%, 1\%, 30\%$, and $100\%$) used during training. 
We show that the proposed boundary based method is superior to the pixel-wise methods in Fig.~\ref{fig:kitti_qual}. 
The baseline method SNE-RoadSeg produces a large number of noisy pixels when few labels are presented, e.g., results in the fourth row of Fig.~\ref{fig:kitti_qual}. The baseline OFF-Net has a similar problem (see the fifth line) although the noise is much less than  SNE-RoadSeg. 
This is because the training data is insufficient and the model is still underfitting to the data, leading to large prediction errors. However,  pixel-wise methods can still show large amounts of noise even when we increase the number of labels in the training data; see the second- and third-to-the-last rows in Fig.~\ref{fig:kitti_qual}. One could argue that the noise appears because the training data is insufficient, and that simply collecting more training data would fix the problem. However,
in practice, it is very difficult to collect
large amounts of training data, and even when
large datasets are available, pixel-wise noise
still easily occurs when we try to deploy a model
in a new environment with a domain shift
from the training data.

\begin{figure} \vspace{7pt}
{
  \centering
    \subfigure[]
{\label{fig:kitti_iou}\includegraphics[width=0.49\linewidth]{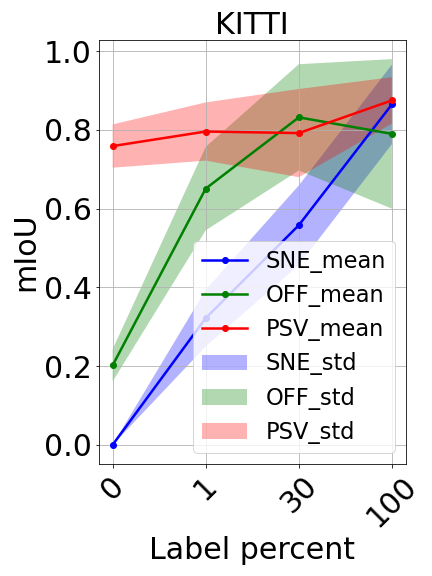}}
  	\subfigure[]
{\label{fig:city_iou}\includegraphics[width=0.49\linewidth]{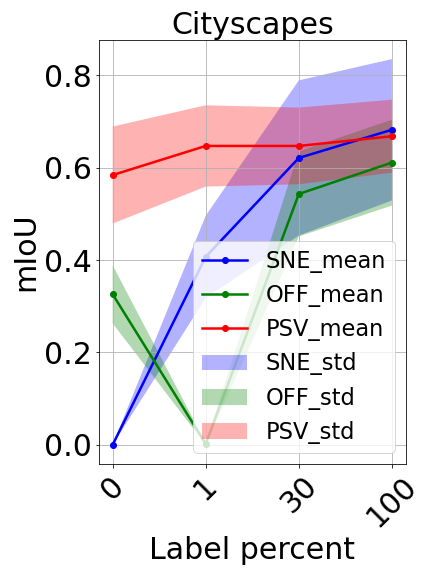}} 
  \caption{\small Quantitative comparisons of different methods as a function of the quantity of ground turth available during training, on (a) KITTI dataset and (b) Cityscapes dataset.
  }
\label{fig:iou_curves}  
}
\end{figure}

In contrast, we  observe that the proposed boundary-based PSV-Net is able to yield estimates (the $3^{rd}$ row) close to the ground truth labels even without any ground truth labels provided during training. All models can achieve better predictions as the amount of ground truth increases, but PSV-Net still maintains an advantage over the baselines. There are few noisy pixels appearing in the resulting predictions, which in turn can significantly improve results of the downstream planning modules. A more detailed quantitative comparison can be seen in Fig.~\ref{fig:kitti_iou}. 

\subsection{Comparison for Cityscapes} 
We show a qualitative comparison for the Cityscapes dataset in Fig. \ref{fig:city_qual}, again comparing results across different amounts of available ground truth labels.
We see that the navigable spaces in Cityscapes exhibit more complicated patterns than KITTI due to the existence of more objects in the scene (e.g., cars and pedestrians), but our proposed method can still detect the shape of the navigable space and outperform the baseline when no or few labels are available. A quantitative comparison is in Fig. \ref{fig:city_iou}. Note that the output of OFF-Net collapses with very few labels ($1\%$), because the model is not able to learn a reasonable representation. Interestingly, the mIoU with 0\% labels is greater than the mIoU with 1\%. One possible explanation is that the 0\% model generates random outputs based on randomly-initialized network weights, which are actually better than the model trained with 1\%.

As the number of ground truth labels increases, the baseline results improve significantly while  PSV-Net remains about the same. The reasons for this are two-fold. First, we do not apply data selection to  Cityscapes, and there exist many images having  inconsistency between  geometry and semantics. While our  method may generate  geometrically-correct segmentation maps that are required for autonomous navigation, it may not match the semantics defined in the ground truth labels, which
counts against our accuracy.
An example of the mismatch between the geometry and semantics (ground truth label) is shown in the last column of Fig. \ref{fig:city_qual}, where the \textit{sidewalk} and \textit{road} share the same geometric plane and thus our segmentation model treats both classes as  navigable space (blue color). However, according to the  ground truth label, only \textit{road} is defined as navigable. 

Second, the number of degrees of freedom of the proposed method is less than that of the baselines. The degrees of freedom here can be interpreted as the dimensionality of the prediction space. Our method  predicts a set of 2D  boundary point coordinates, so the degrees of freedom depends on the number of  points, which is usually a small number --- 30 for KITTI, and 50 for Cityscapes. However, for pixel-wise methods the number of degrees of freedom is the number of pixels in the image, which is usually an enormous number --- $1242\times 375$ for KITTI and $1024\times 512$ for Cityscapes. From the above analysis, we can see that the number of degrees of freedom of the baselines is significantly larger than that of our  method, so they have higher capacity to predict highly complicated
structures.
However, as shown in the baseline predictions in Figs. \ref{fig:kitti_qual} and \ref{fig:city_qual}, the disadvantage of many degrees of freedom is that more noise can be generated, which might confuse planning modules during navigation.

\begin{table} \vspace{7pt}
\caption{\small Comparison of domain adaptive performance for different methods. }
\label{tab:adaptation}
\begin{tabular}{c}
\hline \hline
\\
\hline
$\%$\\
\hline 
0\\
1\\
30\\
100\\
\hline \hline
\end{tabular}
\begin{tabular}{ccc}
\hline \hline
\multicolumn{3}{c}{KITTI $\rightarrow$ ORFD}\\
\hline
\textbf{S}NE&\textbf{O}FF&\textbf{P}SV\\
\hline
0.125$\pm$0.009 & 0.198$\pm$0.023 & 0.762$\pm$0.038\\
0.253$\pm$0.051 & 0.463$\pm$0.092 & 0.781$\pm$0.092\\
0.422$\pm$0.109 & 0.663$\pm$0.102 & 0.788$\pm$0.109\\
0.673$\pm$0.142 & 0.684$\pm$0.173 & 0.851$\pm$0.132\\
\hline \hline
\end{tabular}
\end{table} 

\begin{figure}
{
\centering
  {\includegraphics[width=\linewidth]{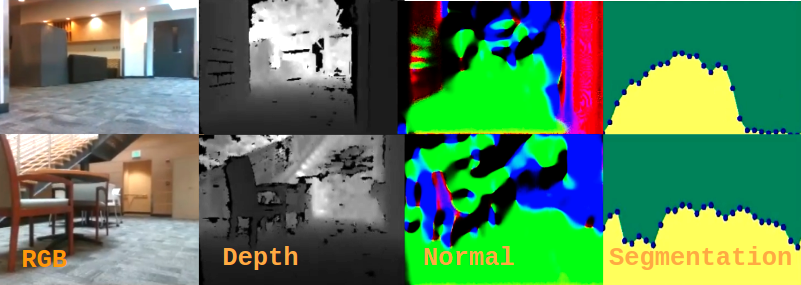}}
\caption{\small Two scenarios of indoor data, each of which is listed as one row.
} 
\label{fig:indoor_data}  
}
\end{figure}

\subsection{Generalization Evaluation}
Generalization performance is crucial for applying deep models in real-world problems. To compare the generalization  of different methods, we conduct an   experiment  in which  all models are  trained on KITTI but  tested  on ORFD, which is a very different dataset.
We report quantitative results (measured with mIoU) in Table \ref{tab:adaptation}. We see that the performance of baselines (SNE and OFF) are significantly degraded at all ground truth label amounts compared to the KITTI testing experiments (see Fig. \ref{fig:kitti_iou}). However, our PSV-Net  still maintains relatively good results.

\subsection{Indoor Experiments and Navigation}

The experiments in the previous sections showed that our proposed PSV-Net can work well even if no ground truth labels are provided. This is a significant advantage when deploying the model to novel environments, since collecting data for new scenarios can be prohibitively expensive. We validate the proposed model on our INDOOR dataset, which has no ground truth labels. Qualitative results are illustrated in Fig. \ref{fig:indoor_data}. The trained model on the INDOOR data can be directly combined with our visual planner to build a navigation system. Thanks to the efficient design of PSV-Net, during inference, we only use the trained $E_2$ to directly predict boundary vertices from input surface normal images. The inference speed is 36 frames per second on average, permitting real-time applications. 
By integrating the proposed PSV-Net with the visual planner, we succeed in building a visual navigation system on a ground robot. We only use an RGB-D camera on the robot as the perception sensor. Navigation demonstrations are provided in the supplementary video (also public video link: \url{https://www.youtube.com/watch?v=LAliV-rUYGE}).

\begin{figure} \vspace{7pt}
{
  \centering
    \subfigure[]
  	{\label{fig:a_1}\includegraphics[width=0.32\linewidth]{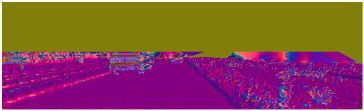}}
  	\subfigure[]
  	{\label{fig:a_2}\includegraphics[width=0.32\linewidth]{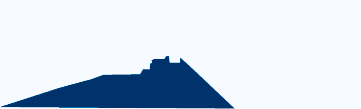}} 
  	\subfigure[]
  	{\label{fig:a_3}\includegraphics[width=0.32\linewidth]{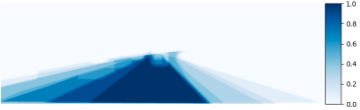}} 
  	\subfigure[]
  	{\label{fig:a_4}\includegraphics[width=0.32\linewidth]{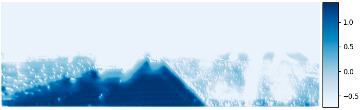}} 
  	\subfigure[]
  	{\label{fig:a_5}\includegraphics[width=0.32\linewidth]{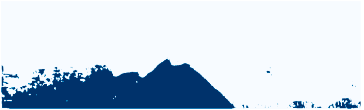}} 
  	\subfigure[]
  	{\label{fig:a_6}\includegraphics[width=0.32\linewidth]{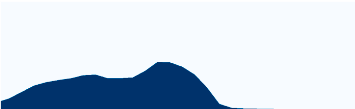}}
  \caption{\small Data flow in PSV-Net. (a) Surface normal image, obtained based on the lidar data provided in KITTI dataset, hence half of values are missing. (b) Ground truth label. (c) Prior for Net-I. (d) Latent representation in Net-I. (e) Pseudo label generated by Net-I. (f) Segmentation output rendered from $D_2$. The blue region indicates  freespace. 
  }
\label{fig:ablation}  
}
\end{figure}

\begin{table}
\caption{\small Quantitative comparison of  Net-I and  Net-II for both KITTI and Cityscapes. The columns starting from I are results for Net-I while II is for Net-II. The columns starting from Gain are advantageous gains of Net-II over Net-I. }
\label{tab:ablation}
\begin{tabular}{c}
\hline \hline
\\
\\
\hline 
\textbf{A}\\
\textbf{P}\\
\textbf{R}\\
\textbf{F}\\
\textbf{I}\\
\hline \hline
\end{tabular}
\quad
\begin{tabular}{cca}
\hline \hline
\multicolumn{3}{c}{KITTI}\\
\hline
I&II&Gain\\
\hline
92.2 & 93.9 & +1.7\\ 
89.3 & 91.0 & +1.7\\
78.4 & 78.5 & +0.1\\
82.1 & 83.6 & +1.5\\
70.7 & 72.1 & +1.4\\
\hline \hline
\end{tabular}
\quad
\begin{tabular}{cca}
\hline \hline
\multicolumn{3}{c}{Cityscapes}\\
\hline
I&II&Gain\\
\hline
82.3 & 84.3 & +2.0\\
73.9 & 74.3 & +0.4\\
72.6 & 75.7 & +3.1\\
72.9 & 73.2 & +0.3\\
57.8 & 58.4 & +0.6\\
\hline \hline
\end{tabular}
\end{table}

\subsection{Ablation Studies} 

We first show the data flow in a fully self-supervised trained PSV-Net and validate the necessity of each module in our network structure (see Fig.~\ref{fig:ablation}) where one image from KITTI dataset is used as an example. We construct the prior using 10 randomly-selected  ground truth labels in KITTI. The goal of specifying a prior is to inform the model that the navigable space is generally in the lower middle part of a first-person view image. Fig.~\ref{fig:a_4} shows the first channel of the latent representation (two channels in total since we only perform binary segmentation) of  Net-I. Fig.~\ref{fig:a_5} is the pixel-wise label prediction results from the latent representation of Net-I. Although Fig.~\ref{fig:a_5} is able to show the shape of the navigable space, there is still abundant noise. We use this noisy prediction as the pseudo label to train  Net-II, and the prediction result (Fig.~\ref{fig:a_6}) is accurate and clean.

To further show the advantage of combining Net-I and Net-II, we quantitatively compare the performance of the two nets for both KITTI and Cityscapes. To perform a thorough ablation study, besides the m\textbf{I}oU (see Section \ref{sec:baselines_and_metrics}), we also use several other metrics, including \textbf{A}ccuracy, \textbf{P}recision, \textbf{R}ecall, \textbf{F}-Score. The definitions for those additional metrics are $\mathbf{A} = \frac{n_{tp}+n_{tn}}{n_{tp} + n_{tn} + n_{fp} + n_{fn}}, \mathbf{P} = \frac{n_{tp}}{n_{tp} + n_{fp}}, \mathbf{R} = \frac{n_{tp}}{n_{tp} + n_{fn}}$, and $\mathbf{F} = \frac{2 n_{tp}^2}{2 n_{tp}^2+n_{tp}(n_{fp} + n_{fn})}$ As shown in Table~\ref{tab:ablation}, the segmentation performance is systematically improved with the refinement of  Net-II. This indicates our proposed polyline-based segmentation is able to reduce the noise in pixel-wise predictions.

\section{Conclusion}
We propose a new framework, PSV-Net, to learn the navigable space in a self-supervised fashion. The proposed framework discretizes the boundary of navigable spaces into a set of vertices. Through extensive evaluations, we have validated the effectiveness of the proposed method and its advantages over the state-of-the-art fully-supervised learning baseline methods. We also validate the effectiveness of the proposed PSV-Net with a visual planning method for efficient autonomous navigation.

\bibliographystyle{IEEEtran}
\bibliography{ref}

\end{document}